\documentclass{article} 
\usepackage{nips15submit_e,times}
\usepackage{hyperref}
\usepackage{url}
\usepackage{graphicx}
\usepackage[numbers]{natbib}

\usepackage{qtree}
\usepackage{ulem}

\usepackage{multicol}

\usepackage{enumitem}

\renewcommand{\#}[1]{\textbf{#1}}

\usepackage{tikz}
\usetikzlibrary{matrix,chains,positioning,decorations.pathreplacing,arrows}

\nipsfinalcopy

\graphicspath{{./figures/}}

\author{
Mengye Ren${}^1$, Ryan Kiros${}^1$, Richard S. Zemel${}^{1, 2}$\\
University of Toronto${}^1$\\
Canadian Institute for Advanced Research${}^2$\\
\texttt{\{mren, rkiros, zemel\}@cs.toronto.edu}
}
\title{Exploring Models and Data for Image Question Answering}

\begin{document}
\maketitle
\begin{abstract}
This work aims to address the problem of image-based question-answering (QA)
with new models and datasets. In our work, we propose to use neural networks
and visual semantic embeddings, without intermediate stages such as object
detection and image segmentation, to predict answers to simple questions about
images. Our model performs 1.8 times better than the only published results on
an existing image QA dataset. We also present a question generation algorithm
that converts image descriptions, which are widely available, into QA form. We
used this algorithm to produce an order-of-magnitude larger dataset, with more
evenly distributed answers. A suite of baseline results on this new dataset are
also presented.
\end{abstract}

\section{Introduction}
Combining image understanding and natural language interaction is one of the
grand dreams of artificial intelligence. We are interested in the problem of
jointly learning image and text through a question-answering task. Recently,
researchers studying image caption generation~\cite{vinyals14,kiros14b,
karpathy14,mao14,donahue14,chen14,fang14,xu15,lebret15,klein15} have developed
powerful methods of jointly learning from image and text inputs to form higher
level representations from models such as convolutional neural networks (CNNs)
trained on object recognition, and word embeddings trained on large scale text
corpora. Image QA involves an extra layer of interaction between human and
computers. Here the model needs to pay attention to details of the image
instead of describing it in a vague sense. The problem also combines many
computer vision sub-problems such as image labeling and object detection.

In this paper we present our contributions to the problem: a generic end-to-end
QA model using visual semantic embeddings to connect a CNN and a recurrent
neural net (RNN), as well as comparisons to a suite of other models; an
automatic question generation algorithm that converts description sentences
into questions; and a new QA dataset (COCO-QA) that was generated using the
algorithm, and a number of baseline results on this new dataset.

In this work we assume that the answers consist of only a single word,
which allows us to treat the problem as a classification problem. This also
makes the evaluation of the models easier and more robust, avoiding the thorny
evaluation issues that plague multi-word generation problems.

\begin{figure}
\centering
\scriptsize
$\begin{array}{p{3.2cm} p{3.2cm} p{3.2cm} p{3.2cm}}

\scalebox{0.23}{
\includegraphics[width=\textwidth, height=.7\textwidth]{1553.jpg}}

\parbox{3.2cm}{
\vskip 0.05in
DAQUAR 1553\\
\textbf{What is there in front of the sofa?}\\
Ground truth: table\\
IMG+BOW: \textcolor{green}{table (0.74)}\\
2-VIS+BLSTM: \textcolor{green}{table (0.88) }\\
LSTM: \textcolor{red}{chair (0.47) }
}
&

\scalebox{0.23}{
\includegraphics[width=\textwidth, height=.7\textwidth]{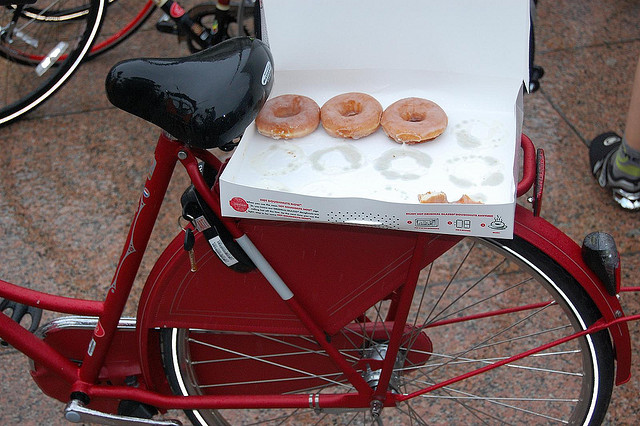}}

\parbox{3.2cm}{
\vskip 0.05in
COCOQA 5078\\
\textbf{How many leftover donuts is the red bicycle holding?}\\
Ground truth: three\\
IMG+BOW: \textcolor{red}{two (0.51)}\\
2-VIS+BLSTM: \textcolor{green}{three (0.27)}\\
BOW: \textcolor{red}{one (0.29)}
}
&

\scalebox{0.23}{
\includegraphics[width=\textwidth, height=.7\textwidth]{1238.jpg}}

\parbox{3.2cm}{
\vskip 0.05in
COCOQA 1238\\
\textbf{What is the color of the tee-shirt?}\\
Ground truth: blue\\
IMG+BOW: \textcolor{green}{blue (0.31) }\\
2-VIS+BLSTM: \textcolor{red}{orange (0.43) }\\
BOW: \textcolor{red}{green (0.38) }
}
&

\scalebox{0.23}{
\includegraphics[width=\textwidth, height=.7\textwidth]{135.jpg}}

\parbox{3.2cm}{
\vskip 0.05in
COCOQA 26088\\
\textbf{Where is the gray cat sitting?}\\
Ground truth: window\\
IMG+BOW: \textcolor{green}{window (0.78) }\\
2-VIS+BLSTM: \textcolor{green}{window (0.68) }\\
BOW: \textcolor{red}{suitcase (0.31) }
}
\end{array}$

\caption{Sample questions and responses of a variety of models. 
Correct answers are in green and incorrect in red. The numbers in parentheses 
are the probabilities assigned to the top-ranked answer by the given model. 
The leftmost example is from the DAQUAR dataset, and the others are from our 
new COCO-QA dataset.}
\label{fig:demo_questions}
\end{figure}

\section{Related Work}
Malinowski and Fritz~\cite{malinowski14a} released a dataset with images and
question-answer pairs, the DAtaset for QUestion Answering on Real-world images
(DAQUAR). All images are from the NYU depth v2 dataset~\cite{silberman12}, and
are taken from indoor scenes. Human segmentation, image depth values, and
object labeling are available in the dataset. The QA data has two sets of
configurations, which differ by the number of object classes appearing in the
questions (37-class and 894-class). There are mainly three types of questions
in this dataset: object type, object color, and number of objects. Some
questions are easy but many questions are very hard to answer even for humans.
Since DAQUAR is the only publicly available image-based QA dataset, it is one
of our benchmarks to evaluate our models.

Together with the release of the DAQUAR dataset, Malinowski and Fritz presented
an approach which combines semantic parsing and image segmentation. Their
approach is notable as one of the first attempts at image QA, but it has a
number of limitations. First, a human-defined possible set of predicates are
very dataset-specific. To obtain the predicates, their algorithm also depends
on the accuracy of the image segmentation algorithm and image depth
information.  Second, their model needs to compute all possible spatial
relations in the training images. Even though the model limits this to the
nearest neighbors of the test images, it could still be an expensive operation
in larger datasets.  Lastly the accuracy of their model is not very strong. We
show below that some simple baselines perform better.

Very recently there has been a number of parallel efforts on both creating
datasets and proposing new models~\cite{antol14, malinowski15, gao15, ma15}.
Both Antol et al.~\cite{antol14} and Gao et al.~\cite{gao15} used MS-COCO
\cite{mscoco} images and created an open domain dataset with human generated
questions and answers. In Anto et al.'s work, the authors also included cartoon
pictures besides real images. Some questions require logical reasoning in order
to answer correctly. Both Malinowski et al.~\cite{malinowski15} and Gao et al.
\cite{gao15} use recurrent networks to encode the sentence and output the
answer. Whereas Malinowski et al. use a single network to handle both encoding
and decoding, Gao et al. used two networks, a separate encoder and decoder.
Lastly, bilingual (Chinese and English) versions of the QA dataset are
available in Gao et al.'s work. Ma et al.~\cite{ma15} use CNNs to both extract
image features and sentence features, and fuse the features together with
another multi-modal CNN.

Our approach is developed independently from the work above. Similar to the
work of Malinowski et al. and Gao et al., we also experimented with recurrent
networks to consume the sequential question input. Unlike Gao et al., we
formulate the task as a classification problem, as there is no single well-
accepted metric to evaluate sentence-form answer
accuracy~\cite{mscoco_captions}. Thus, we place more focus on a limited domain
of questions that can be answered with one word. We also formulate and evaluate
a range of other algorithms, that utilize various representations drawn from
the question and image, on these datasets.

\section{Proposed Methodology}
The methodology presented here is two-fold. On the model side we develop and
apply various forms of neural networks and visual-semantic embeddings on this
task, and on the dataset side we propose new ways of synthesizing QA pairs from
currently available image description datasets.

\subsection{Models}
In recent years, recurrent neural networks (RNNs) have enjoyed some successes
in the field of natural language processing (NLP). Long short-term memory
(LSTM)~\cite{hochreiter97} is a form of RNN which is easier to train than
standard RNNs because of its linear error propagation and multiplicative
gatings. Our model builds directly on top of the LSTM sentence model and is
called the ``VIS+LSTM'' model. It treats the image as one word of the question.
We borrowed this idea of treating the image as a word from caption generation
work done by Vinyals et al.~\cite{vinyals14}. We compare this newly
proposed model with a suite of simpler models in the Experimental Results
section.

\begin{enumerate}
\item We use the last hidden layer of the 19-layer Oxford VGG Conv Net
\cite{simonyan14} trained on ImageNet 2014 Challenge~\cite{ilsvrc14} as our
visual embeddings. The CNN part of our model is kept frozen during training.

\item We experimented with several different word embedding models: randomly
initialized embedding, dataset-specific skip-gram embedding and general-purpose
skip-gram embedding model~\cite{mikolov13}. The word embeddings are trained
with the rest of the model.

\item We then treat the image as if it is the first word of the sentence.
Similar to DeViSE~\cite{frome13}, we use a linear or affine transformation to
map 4096 dimension image feature vectors to a 300 or 500 dimensional vector
that matches the dimension of the word embeddings.

\item We can optionally treat the image as the last word of the question as
well through a different weight matrix and optionally add a reverse LSTM, which
gets the same content but operates in a backward sequential fashion.

\item The LSTM(s) outputs are fed into a softmax layer at the last timestep to
generate answers.
\end{enumerate}

\begin{figure}
\centering
\scalebox{0.6}{
\begin{tikzpicture}[scale=0.9]
\Large
\foreach \x in  {5, 8, 11, 17}
{
	\draw (\x,1) rectangle (\x+2,2);
	\draw [-latex](\x+1,0) -- (\x+1,1);
}

\foreach \x in  {8, 11, 17}
{
	\draw [-latex](\x+1,-1.5) -- (\x+1,-1);
}

\foreach \x in  {8, 11, 14, 17}
{
	\draw [-latex](\x-1,1.5) -- (\x,1.5);
}

\path (9,-2.5) node[draw=none] (t0) {$t=1$};
\path (12,-2.5) node[draw=none] (t1) {$t=2$};
\path (18,-2.5) node[draw=none] (tT1) {$t=T$};
\draw [dotted] (14,1.5) -- (17,1.5);

\path (9,-1.75) node[draw=none] (w1) {``How''};
\path (12,-1.75) node[draw=none] (w2) {``many''};
\path (18,-1.75) node[draw=none] (w3) {``books''};
\path (6,1.5) node[draw=none] (lstm1) {LSTM};

\draw (8,3) rectangle (19,4);
\draw(12,3.5) circle (0.2cm);
\draw [fill=red] (13.5,3.5) circle (0.2cm);
\path (15,3.5) node[draw=none] {...};
\draw (16.5,3.5) circle (0.2cm);
\draw (18,3.5) circle (0.2cm);
\path (10,3.5) node[draw=none] {Softmax};
\path (12,4.5) node[draw=none] {One};
\path (13.5,4.5) node[draw=none] {Two};
\path (15,4.5) node[draw=none] {...};
\path (16.5,4.5) node[draw=none] {Red};
\path (18,4.5) node[draw=none] {Bird};
\path (12,5.25) node[draw=none] {.21};
\path (13.5,5.25) node[draw=none] {.56};
\path (15,5.25) node[draw=none] {...};
\path (16.5,5.25) node[draw=none] {.09};
\path (18,5.25) node[draw=none] {.01};
\draw [-latex] (18,2) -- (18,3);

\draw (5,-1) rectangle (7, 0);
\path (6,-0.5) node[draw=none] (imgMap) {Linear};

\draw (1.6, -1) rectangle (2.0, 0);
\draw[fill=white] (2.5, -0.5) rectangle (3.0, 0);
\draw[fill=white] (2.3, -0.75) rectangle (2.8, -0.25);
\draw[fill=white] (2.1, -1) rectangle (2.6, -0.6);
\draw (3.1, -1) rectangle (3.5, 0);
\draw (3.6, -1) rectangle (4, 0);

\draw [-latex] (4, -0.5) -- (5, -0.5);
\draw [-latex] (1,-0.5) -- (1.6, -0.5);

\path (0.2,-0.5) node[draw=none] (image) {Image};
\path (3,-1.75) node[draw=none] (cnn) {CNN};

\draw (8, -1) rectangle (19, 0);
\path (13.5,-0.5) node[draw=none] (Wemb) {Word Embedding};

\end{tikzpicture}}
\caption{VIS+LSTM Model}
\label{fig:imgword}
\end{figure}
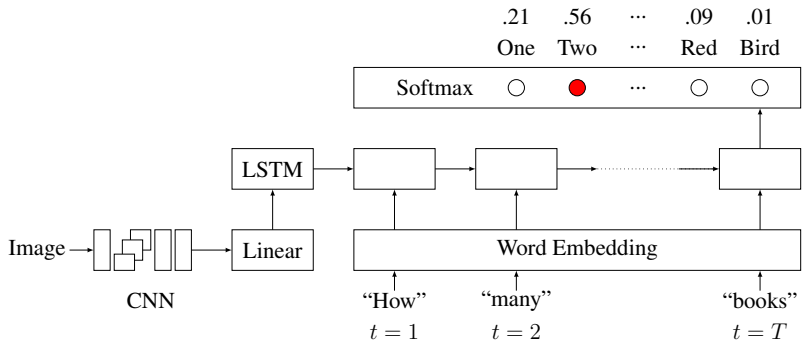

\subsection{Question-Answer Generation}
The currently available DAQUAR dataset contains approximately 1500 images and
7000 questions on 37 common object classes, which might be not enough for
training large complex models. Another problem with the current dataset is that
simply guessing the modes can yield very good accuracy.

We aim to create another dataset, to produce a much larger number of QA pairs
and a more even distribution of answers. While collecting human generated QA
pairs is one possible approach, and another is to synthesize questions based on
image labeling, we instead propose to automatically convert descriptions into
QA form. In general, objects mentioned in image descriptions are easier to
detect than the ones in DAQUAR's human generated questions, and than the ones
in synthetic QAs based on ground truth labeling. This allows the model to
rely more on rough image understanding without any logical reasoning. Lastly
the conversion process preserves the language variability in the original
description, and results in more human-like questions than questions generated
from image labeling.

As a starting point we used
the MS-COCO dataset~\cite{mscoco}, but the same method can be applied to any
other image description dataset, such as Flickr~\cite{flickr8k}, SBU
\cite{ordonez11}, or even the internet.

\subsubsection{Pre-Processing \& Common Strategies}
We used the Stanford parser~\cite{klein03} to obtain the syntatic structure of
the original image description. We also utilized these strategies for forming
the questions.

\begin{enumerate}[leftmargin=*]
\item Compound sentences to simple sentences \\ Here we only consider a simple
case, where two sentences are joined together with a conjunctive word. We split
the orginial sentences into two independent sentences.

\item Indefinite determiners ``a(n)'' to definite determiners ``the''.

\item Wh-movement constraints \\ In English, questions tend to start with
interrogative words such as ``what''. The algorithm needs to move the verb as
well as the ``wh-'' constituent to the front of the sentence. For example: ``A
man is riding a \textbf{horse}'' becomes   ``\textbf{What} is \textbf{the} man
riding?'' In this work we consider the following two simple constraints: (1)
A-over-A principle which restricts the movement of a wh-word inside a noun
phrase (NP)~\cite{chomsky73}; (2) Our algorithm does not move any wh-word that
is contained in a clause constituent.
\end{enumerate}

\subsubsection{Question Generation}
Question generation is still an open-ended topic. Overall, we adopt a conservative 
approach to generating questions in an attempt to create high-quality questions. We 
consider generating four types of questions below:

\begin{enumerate}[leftmargin=*]
\item {\it Object} Questions: First, we consider asking about an
object using ``what''. This involves replacing the actual object with a
``what'' in the sentence, and then transforming the sentence structure so that
the ``what'' appears in the front of the sentence. The entire algorithm has the
following stages: (1) Split long sentences into simple sentences; (2) Change
indefinite determiners to definite determiners; (3) Traverse the sentence and
identify potential answers and replace with ``what''. During the traversal of
object-type question generation, we currently ignore all the prepositional
phrase (PP) constituents; (4) Perform wh-movement. In order to identify a
possible answer word, we used WordNet~\cite{wordnet} and the NLTK software
package~\cite{nltk} to get noun categories.

\item {\it Number} Questions: We follow a similar procedure as the
previous algorithm, except for a different way to identify potential answers:
we extract numbers from original sentences. Splitting compound sentences,
changing determiners, and wh-movement parts remain the same.

\item {\it Color} Questions: Color questions are much easier to
generate. This only requires locating the color adjective and the noun to which
the adjective attaches. Then it simply forms a sentence ``What is the color of
the [object]'' with the ``object'' replaced by the actual noun.

\item {\it Location} Questions: These are similar to generating object
questions, except that now the answer traversal will only search within PP
constituents that start with the preposition ``in''. We also added rules to
filter out clothing so that the answers will mostly be places, scenes, or large
objects that contain smaller objects.
\end{enumerate}

\subsubsection{Post-Processing}
We rejected the answers that appear too rarely or too often in our generated
dataset. After this QA rejection process, the frequency of the most common answer 
words was reduced from 24.98\% down to 7.30\% in the test set of COCO-QA.

\section{Experimental Results}
\subsection{Datasets}
Table~\ref{tab:dataset_category_stats} summarizes the statistics of COCO-QA. It
should be noted that since we applied the QA pair rejection process,
mode-guessing performs very poorly on COCO-QA. However, COCO-QA questions are
actually easier to answer than DAQUAR from a human point of view. This
encourages the model to exploit salient object relations instead of
exhaustively searching all possible relations. COCO-QA dataset can be
downloaded at \url{http://www.cs.toronto.edu/~mren/imageqa/data/cocoqa}

Here we provide some brief statistics of the new dataset. The maximum question
length is 55, and average is 9.65.  The most common answers are ``two'' (3116,
2.65\%), ``white'' (2851, 2.42\%), and ``red'' (2443, 2.08\%). The least common
are ``eagle'' (25, 0.02\%) ``tram'' (25, 0.02\%), and ``sofa'' (25, 0.02\%).
The median answer is ``bed'' (867, 0.737\%).  Across the entire test set (38,948
QAs), 9072 (23.29\%) overlap in training questions, and 7284 (18.70\%) overlap
in training question-answer pairs.
\begin{table}
\caption{COCO-QA question type break-down}
\label{tab:dataset_category_stats}
\vskip 0.15in
\begin{center}
\begin{small}
\begin{sc}
\begin{tabular}{c c c c c}
\hline
Category & Train & \%       & Test  & \%       \\
\hline
Object   & 54992 & 69.84\%  & 27206 & 69.85\%  \\
Number   & 5885  & 7.47\%   & 2755  & 7.07\%   \\
Color    & 13059 & 16.59\%  & 6509  & 16.71\%  \\
Location & 4800  & 6.10\%   & 2478  & 6.36\%   \\
\hline
Total    & 78736 & 100.00\% & 38948 & 100.00\% \\
\hline
\end{tabular}
\end{sc}
\end{small}
\end{center}
\end{table}

\subsection{Model Details}
\begin{enumerate}[leftmargin=*]
\item \#{VIS+LSTM}: The first model is the CNN and LSTM with a
dimensionality-reduction weight matrix in the middle; we call this ``VIS+LSTM''
in our tables and figures.

\item \#{2-VIS+BLSTM}: The second model has two image feature inputs, at the
start and the end of the sentence, with different learned linear
transformations, and also has LSTMs going in both the forward and backward
directions. Both LSTMs output to the softmax layer at the last timestep. We
call the second model ``2-VIS+BLSTM''.

\item \#{IMG+BOW}: This simple model performs multinomial logistic regression
based on the image features without dimensionality reduction (4096 dimension),
and a bag-of-word (BOW) vector obtained by summing all the learned word vectors
of the question.

\item \#{FULL}: Lastly, the ``FULL'' model is a simple average of the three
models above.
\end{enumerate}

We release the complete details of the models at
\url{https://github.com/renmengye/imageqa-public}.

\subsection{Baselines}
To evaluate the effectiveness of our models, we designed a few baselines.

\begin{enumerate}[leftmargin=*]
\item \#{GUESS}: One very simple baseline is to predict the mode based on the
question type. For example, if the question contains ``how many'' then the
model will output ``two.'' In DAQUAR, the modes are ``table'', ``two'', and
``white'' and in COCO-QA, the modes are ``cat'', ``two'', ``white'', and
``room''.

\item \#{BOW}: We designed a set of ``blind'' models which are given only the
questions without the images. One of the simplest blind models performs
logistic regression on the BOW vector to classify answers.

\item \#{LSTM}: Another ``blind'' model we experimented with simply inputs the
question words into the LSTM alone.

\item \#{IMG}: We also trained a counterpart ``deaf'' model. For each type of
question, we train a separate CNN classification layer (with all lower layers
frozen during training). Note that this model knows the type of question, in
order to make its performance somewhat comparable to models that can take
into account the words to narrow down the answer space. However the model does
not know anything about the question except the type.

\item \#{IMG+PRIOR}: This baseline combines the prior knowledge of an object
and the image understanding from the ``deaf model''. For example, a question
asking the color of a white bird flying in the blue sky may output white rather
than blue simply because the prior probability of the bird being blue is lower.
We denote $c$ as the color, $o$ as the class of the object of interest, and $x$
as the image. Assuming $o$ and $x$ are conditionally independent given the
color,
\begin{equation} 
p(c | o, x)
= \frac{p(c, o | x)}{\sum_{c \in \mathcal{C}} p(c, o | x)} 
= \frac{p(o | c, x) p(c | x)}{\sum_{c \in \mathcal{C}} p(o | c, x) p(c | x)}
= \frac{p(o | c) p(c | x)}{\sum_{c \in \mathcal{C}}p(o | c) p(c | x)}
\end{equation}

This can be computed if $p(c | x)$ is the output of a logistic regression given
the CNN features alone, and we simply estimate $p(o | c)$ empirically:
$\hat{p}(o | c) = \frac{count(o, c)}{count(c)}$. We use Laplace smoothing on
this empirical distribution.

\item \#{K-NN}: In the task of image caption generation, Devlin et al.
\cite{devlin15} showed that a nearest neighbors baseline approach actually
performs very well. To see whether our model memorizes the training data for
answering new question, we include a K-NN baseline in the results. Unlike
image caption generation, here the similarity measure includes both image
and text.  We use the bag-of-words representation learned from IMG+BOW, and
append it to the CNN image features.  We use Euclidean distance as the
similarity metric; it is possible to improve the nearest neighbor result by 
learning a similarity metric.
\end{enumerate}

\subsection{Performance Metrics}
To evaluate model performance, we used the plain answer accuracy as well as the
Wu-Palmer similarity (WUPS) measure~\cite{wu94, malinowski14b}. The WUPS
calculates the similarity between two words based on their longest common
subsequence in the taxonomy tree. If the similarity between two words is less
than a threshold then a score of zero will be given to the candidate answer.
Following Malinowski and Fritz~\cite{malinowski14b}, we measure all models
in terms of accuracy, WUPS 0.9, and WUPS 0.0.

\subsection{Results and Analysis}
Table~\ref{tab:daquar_results} summarizes the learning results on DAQUAR
and COCO-QA. For DAQUAR
we compare our results with~\cite{malinowski14b} and~\cite{malinowski15}. It
should be noted that our DAQUAR results are for the portion of the dataset (98.3\%)
with single-word answers. After the release of our paper, Ma et al. \cite{ma15}
claimed to achieve better results on both datasets.

\begin{table}[h]
\caption{DAQUAR and COCO-QA results}
\label{tab:daquar_results}
\begin{center}
\begin{small}
\begin{sc}
\begin{tabular}{c | c c c | c c c}
\hline     
                     &          &  DAQUAR  &          &          &  COCO-QA &          \\
                     & Acc.     & WUPS 0.9 & WUPS 0.0 & Acc.     & WUPS 0.9 & WUPS 0.0 \\
\hline         
MULTI-WORLD
\cite{malinowski14b} &  0.1273  &  0.1810  &  0.5147  &  -       &  -       &  -       \\
GUESS                &  0.1824  &  0.2965  &  0.7759  &  0.0730  &  0.1837  &  0.7413  \\
BOW                  &  0.3267  &  0.4319  &  0.8130  &  0.3752  &  0.4854  &  0.8278  \\
LSTM                 &  0.3273  &  0.4350  &  0.8162  &  0.3676  &  0.4758  &  0.8234  \\
IMG                  &  -       &  -       &  -       &  0.4302  &  0.5864  &  0.8585  \\
IMG+PRIOR            &  -       &  -       &  -       &  0.4466  &  0.6020  &  0.8624  \\
K-NN (K=31, 13)      &  0.3185  &  0.4242  &  0.8063  &  0.4496  &  0.5698  &  0.8557  \\
IMG+BOW              &  0.3417  &  0.4499  &  0.8148  &\#{0.5592}&\#{0.6678}&\#{0.8899}\\
VIS+LSTM             &  0.3441  &  0.4605  &\#{0.8223}&  0.5331  &  0.6391  & 0.8825   \\
ASK-NEURON
\cite{malinowski15}  &  0.3468  &  0.4076  & 0.7954   &  -       &  -       &  -       \\
2-VIS+BLSTM          &\#{0.3578}&\#{0.4683}& 0.8215   &  0.5509  &  0.6534  & 0.8864   \\
FULL                 &\#{0.3694}&\#{0.4815}&\#{0.8268}&\#{0.5784}&\#{0.6790}&\#{0.8952}\\
\hline
HUMAN                &  0.6027  &  0.6104  &  0.7896  &  -       &  -       &  -       \\
\hline
\end{tabular}
\end{sc}
\end{small}
\end{center}
\end{table}

\begin{table}[h]
\caption{COCO-QA accuracy per category}
\label{tab:cocoqa_acc_breakdown}
\begin{center}
\begin{small}
\begin{sc}
\begin{tabular}{c c c c c}
\hline
           &   Object  &   Number &   Color   & Location \\
\hline
GUESS      &   0.0239  &   0.3606 &   0.1457  & 0.0908   \\
BOW        &   0.3727  &   0.4356 &   0.3475  & 0.4084   \\
LSTM       &   0.3587  &   0.4534 &   0.3626  & 0.3842   \\
IMG        &   0.4073  &   0.2926 &   0.4268  & 0.4419   \\
IMG+PRIOR  &   -       &   0.3739 &   0.4899  & 0.4451   \\
K-NN       &   0.4799  &   0.3699 &   0.3723  & 0.4080   \\
IMG+BOW    &\#{0.5866} &   0.4410 &\#{0.5196} &\#{0.4939}\\
VIS+LSTM   &   0.5653  &\#{0.4610}&   0.4587  & 0.4552   \\
2-VIS+BLSTM&   0.5817  &   0.4479 &   0.4953  & 0.4734   \\
FULL       &\#{0.6108} &\#{0.4766}&   0.5148  &\#{0.5028}\\
\hline
\end{tabular}
\end{sc}
\end{small}
\end{center}
\end{table}

From the above results we observe that our model outperforms the baselines and
the existing approach in terms of answer accuracy and WUPS. Our VIS+LSTM and
Malinkowski et al.'s recurrent neural network model~\cite{malinowski15} achieved
somewhat similar performance on DAQUAR. A simple average of all three models
further boosts the performance by 1-2\%, outperforming other models.

It is surprising to see that the IMG+BOW model is very strong on both datasets. One
limitation of our VIS+LSTM model is that we are not able to consume image
features as large as 4096 dimensions at one time step, so the dimensionality
reduction may lose some useful information. We tried to give IMG+BOW a 500 dim.
image vector, and it does worse than VIS+LSTM ($\approx$48\%). 

By comparing the blind versions of the BOW and LSTM models, we hypothesize
that in Image QA tasks, and in particular on the simple questions studied here,
sequential word interaction may not be as important as in other natural
language tasks. 

It is also interesting that the blind model does not lose much on the DAQUAR
dataset, We speculate that it is likely that the ImageNet images are very
different from the indoor scene images, which are mostly composed of furniture.
However, the non-blind models outperform the blind models by a large margin on
COCO-QA. There are three possible reasons: (1) the objects in MS-COCO
resemble the ones in ImageNet more; (2) MS-COCO images have fewer objects
whereas the indoor scenes have considerable clutter; and (3) COCO-QA has more
data to train complex models.

\begin{figure}
\centering\tiny
$\begin{array}{p{3.2cm} p{3.2cm} p{3.2cm} p{3.2cm}}

\scalebox{0.23}{
\includegraphics[width=\textwidth, height=.7\textwidth]{33827.jpg}}

\parbox{3.2cm}{
\vskip 0.05in
COCOQA 33827\\
\textbf{What is the color of the cat?}\\
Ground truth: black\\
IMG+BOW: \textcolor{green}{black (0.55)}\\
2-VIS+LSTM: \textcolor{green}{black (0.73)}\\
BOW: \textcolor{red}{gray (0.40)}

\vskip 0.05in
COCOQA 33827a\\
\textbf{What is the color of the couch?}\\
Ground truth: red\\
IMG+BOW: \textcolor{green}{red (0.65)}\\
2-VIS+LSTM: \textcolor{red}{black (0.44)}\\
BOW: \textcolor{green}{red (0.39)}
}
&

\scalebox{0.23}{
\includegraphics[width=\textwidth, height=.7\textwidth]{1520.jpg}}

\parbox{3.2cm}{
\vskip 0.05in
DAQUAR 1522\\
\textbf{How many chairs are there?}\\
Ground truth: two\\
IMG+BOW: \textcolor{red}{four (0.24)}\\
2-VIS+BLSTM: \textcolor{red}{one (0.29)}\\
LSTM: \textcolor{red}{four (0.19)}

\vskip 0.05in
DAQUAR 1520\\
\textbf{How many shelves are there?}\\
Ground truth: three\\
IMG+BOW: \textcolor{green}{three (0.25)}\\
2-VIS+BLSTM: \textcolor{red}{two (0.48)}\\
LSTM: \textcolor{red}{two (0.21)}
}
&

\scalebox{0.23}{
\includegraphics[width=\textwidth, height=.7\textwidth]{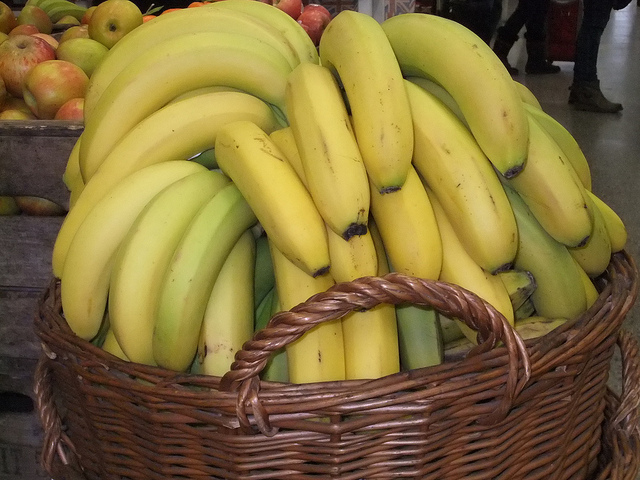}}

\parbox{3.2cm}{
\vskip 0.05in
COCOQA 14855\\
\textbf{Where are the ripe bananas sitting?}\\
Ground truth: basket\\
IMG+BOW: \textcolor{green}{basket (0.97)}\\
2-VIS+BLSTM: \textcolor{green}{basket (0.58)}\\
BOW: \textcolor{red}{bowl (0.48)}

\vskip 0.05in
COCOQA 14855a\\
\textbf{What are in the basket?}\\
Ground truth: bananas\\
IMG+BOW: \textcolor{green}{bananas (0.98) }\\
2-VIS+BLSTM: \textcolor{green}{bananas (0.68)}\\
BOW: \textcolor{green}{bananas (0.14)}
}
&

\scalebox{0.23}{
\includegraphics[width=\textwidth, height=.7\textwidth]{585.jpg}}

\parbox{3.2cm}{
\vskip 0.05in
DAQUAR 585\\
\textbf{What is the object on the chair?}\\
Ground truth: pillow\\
IMG+BOW: \textcolor{red}{clothes (0.37)}\\
2-VIS+BLSTM: \textcolor{green}{pillow (0.65) }\\
LSTM: \textcolor{red}{clothes (0.40)}

\vskip 0.05in
DAQUAR 585a\\
\textbf{Where is the pillow found?}\\
Ground truth: chair\\
IMG+BOW: \textcolor{red}{bed (0.13)}\\
2-VIS+BLSTM: \textcolor{green}{chair (0.17) }\\
LSTM: \textcolor{red}{cabinet (0.79)}
}

\end{array}$
\caption{Sample questions and responses of our system}
\label{fig:selected_questions}
\end{figure}

There are many interesting examples but due to space limitations we can only
show a few in Figure~\ref{fig:demo_questions} and
Figure~\ref{fig:selected_questions}; full results are available at
\url{http://www.cs.toronto.edu/~mren/imageqa/results}.  For some of the
images, we added some extra questions (the ones have an ``a'' in the
question ID); these provide more insight into a model's representation of the
image and question information, and help elucidate questions that our models
may accidentally get correct. The parentheses in the figures represent the
confidence score given by the softmax layer of the respective model.

\textbf{Model Selection}: We did not find that using different word embedding
has a significant impact on the final classification results. We observed that
fine-tuning the word embedding results in better performance and normalizing
the CNN hidden image features into zero-mean and unit-variance helps achieve
faster training time. The bidirectional LSTM model can further boost the result
by a little.
 
\textbf{Object Questions}: As the original CNN was trained for the ImageNet
challenge, the IMG+BOW benefited significantly from its single object recognition
ability. However, the challenging part is to consider spatial relations between
multiple objects and to focus on details of the image. Our models only did a
moderately acceptable job on this; see for instance the first picture of
Figure~\ref{fig:demo_questions} and the fourth picture of
Figure~\ref{fig:selected_questions}. Sometimes a model fails to make a correct
decision but outputs the most salient object, while sometimes the blind model
can equally guess the most probable objects based on the question alone (e.g.,
chairs should be around the dining table). Nonetheless, the FULL model improves
accuracy by 50\% compared to IMG model, which shows the difference between pure
object classification and image question answering.

\textbf{Counting}: In DAQUAR, we could not observe any advantage in the
counting ability of the IMG+BOW and the VIS+LSTM model compared to the blind
baselines. In COCO-QA there is some observable counting ability in very clean
images with a single object type. The models can sometimes count up to five or
six. However, as shown in the second picture of
Figure~\ref{fig:selected_questions}, the ability is fairly weak as they do not
count correctly when different object types are present. There is a lot of room
for improvement in the counting task, and in fact this could be a separate
computer vision problem on its own.

\textbf{Color}: In COCO-QA there is a significant win for the IMG+BOW and the
VIS+LSTM against the blind ones on color-type questions. We further discovered
that these models are not only able to recognize the dominant color of the image
but sometimes associate different colors to different objects, as shown in the
first picture of Figure~\ref{fig:selected_questions}. However, they still
fail on a number of easy examples. Adding prior knowledge provides an
immediate gain on the IMG model in terms of accuracy on Color and Number
questions. The gap between the IMG+PRIOR and IMG+BOW shows some localized color
association ability in the CNN image representation.

\section{Conclusion and Current Directions}

In this paper, we consider the image QA problem and present our end-to-end
neural network models. Our model shows a reasonable understanding of the
question and some coarse image understanding, but it is still very na\"{i}ve in
many situations. While recurrent networks are becoming a popular choice for
learning image and text, we showed that a simple bag-of-words can perform equally
well compared to a recurrent network that is borrowed from an image caption
generation framework~\cite{vinyals14}. We proposed a more complete set of
baselines which can provide potential insight for developing more sophisticated
end-to-end image question answering systems. As the currently available dataset
is not large enough, we developed an algorithm that helps us collect large scale
image QA dataset from image descriptions. Our question generation algorithm is
extensible to many image description datasets and can be automated without
requiring extensive human effort. We hope that the release of the new dataset
will encourage more data-driven approaches to this problem in the future.

Image question answering is a fairly new research topic, and the approach we
present here has a number of limitations. First, our models are just answer
classifiers. Ideally we would like to permit longer answers which will involve
some sophisticated text generation model or structured output. But this will
require an automatic free-form answer evaluation metric. Second, we are only
focusing on a limited domain of questions. However, this limited range of
questions allow us to study the results more in depth. Lastly, it is also hard
to interpret why the models output a certain answer. By comparing our models with
some baselines we can roughly infer whether they understood the image.
Visual attention is another future direction, which could both improve the
results (based on recent successes in image captioning~\cite{xu15}) as well as
help explain the model prediction by examining the attention output at every
timestep.

\subsubsection*{Acknowledgments}

We would like to thank Nitish Srivastava for the support of Toronto Conv Net,
from which we extracted the CNN image features. We would also like to thank
anonymous reviewers for their valuable and helpful comments.

\begin{small}
\bibliography{references}
\bibliographystyle{ieee}
\end{small}

\appendix
\section{Supplementary Material}
\subsection{Question Generation: Syntax Tree Example}
\begin{figure}[h!]
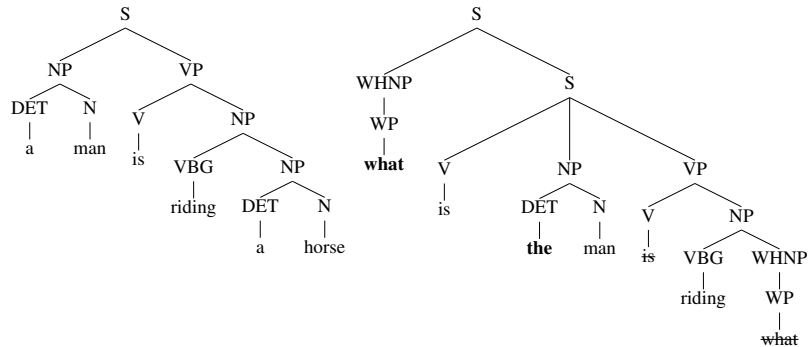

\centering
\small
\scalebox{0.8}{
\Tree [.S [.NP [.DET a ] [.N man ] ] [.VP [.V is ] [.NP [.VBG riding ] [.NP
[.DET a ] [.N horse ] ] ] ] ]} \scalebox{0.8}{ \Tree [.S [.WHNP [.WP
\textbf{what} ] ] [.S [.V is ] [.NP [.DET \textbf{the} ] [.N man ] ] [.VP
[.V \sout{is} ] [.NP [.VBG riding ] [.WHNP [.WP \sout{what} ] ] ] ] ] ]}
\vspace{5mm}
\caption{Example: ``A man is riding a \textbf{horse}'' $=>$ ``\textbf{What} is 
\textbf{the} man riding?''}
\label{fig:what_gen}
\end{figure}

\subsection{Post-Processing of COCO-QA Detail}
\label{app:post_process}
First, answers that appear less than a frequency threshold are discarded.
Second, we enroll a QA pair one at a time.  The probability of enrolling the
next QA pair $(q, a)$ is:
\begin{equation}
p(q, a) = \left\{ \begin{array}{cl}
1 &\mbox{ if count$(a)$ $\le K$ } \\
\exp\left(-\frac{count(a) - K}{[K_2}\right) &\mbox{ otherwise }
\end{array} \right.
\end{equation}
where $count(a)$ denotes the current number of enrolled QA pairs that have $a$
as the ground truth answer, and $K$, $K_2$ are some constants with $K \le K_2$.
In the COCO-QA generation we chose $K = 100$ and $K_2 = 200$. 

\clearpage
\subsection{More Sample Questions and Responses}
\label{sec:more_examples}
\begin{figure}[h]
\centering
\tiny
$\begin{array}{p{3.2cm} p{3.2cm} p{3.2cm} p{3.2cm}}
\scalebox{0.23}{
\includegraphics[width=\textwidth, height=.7\textwidth]{24952.jpg}}

\parbox{3.2cm}{
\vskip 0.05in
COCOQA 23419\\
\textbf{What is the black and white cat wearing?}\\
Ground truth: hat\\
IMG+BOW: \textcolor{green}{hat (0.50)}\\
2-VIS+BLSTM: \textcolor{red}{tie (0.34)}\\
BOW: \textcolor{red}{tie (0.60)}
\vskip 0.05in
COCOQA 23419a\\
\textbf{What is wearing a hat?}\\
Ground truth: cat\\
IMG+BOW: \textcolor{green}{cat (0.94)}\\
2-VIS+BLSTM: \textcolor{green}{cat (0.90)}\\
BOW: \textcolor{red}{dog (0.42)}
}
&

\scalebox{0.23}{
\includegraphics[width=\textwidth, height=.7\textwidth]{2136.jpg}}

\parbox{3.2cm}{
\vskip 0.05in
DAQUAR 2136\\
\textbf{What is right of table?}\\
Ground truth: shelves\\
IMG+BOW: \textcolor{green}{shelves (0.33)}\\
2-VIS+BLSTM: \textcolor{green}{shelves (0.28)}\\
LSTM: \textcolor{green}{shelves (0.20)}

\vskip 0.05in
DAQUAR 2136a\\
\textbf{What is in front of table?}\\
Ground truth: chair\\
IMG+BOW: \textcolor{green}{chair (0.64)}\\
2-VIS+BLSTM: \textcolor{green}{chair (0.31)}\\
LSTM: \textcolor{green}{chair (0.37)}
}
&

\scalebox{0.23}{
\includegraphics[width=\textwidth, height=.7\textwidth]{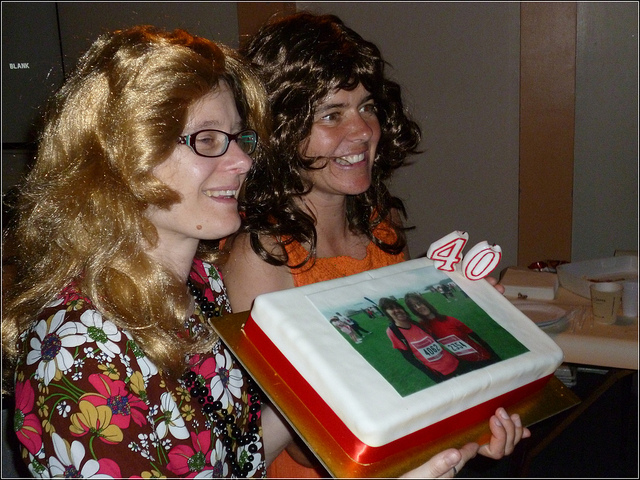}}

\parbox{3.2cm}{
\vskip 0.05in
COCOQA 11372\\
\textbf{What do two women hold with a picture on it?}\\
Ground truth: cake\\
IMG+BOW: \textcolor{green}{cake (0.19)}\\
2-VIS+BLSTM: \textcolor{green}{cake (0.19)}\\
BOW: \textcolor{red}{umbrella (0.15)}
}
&

\scalebox{0.23}{
\includegraphics[width=\textwidth, height=.7\textwidth]{3018.jpg}}

\parbox{3.2cm}{
\vskip 0.05in
DAQUAR 3018\\
\textbf{What is on the right side?}\\
Ground truth: table\\
IMG+BOW: \textcolor{red}{tv (0.28)}\\
2-VIS+LSTM: \textcolor{red}{sofa (0.17)}\\
LSTM: \textcolor{red}{cabinet (0.22)}
}
\\
\noalign{\smallskip}\noalign{\smallskip}

\scalebox{0.23}{
\includegraphics[width=\textwidth, height=.7\textwidth]{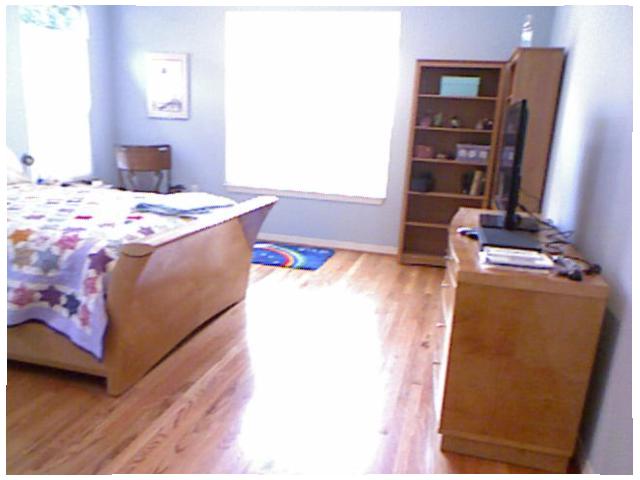}}

\parbox{3.2cm}{
\vskip 0.05in
DAQUAR 1426\\
\textbf{What is on the right side table?}\\
Ground truth: tv\\
IMG+BOW: \textcolor{green}{tv (0.25)}\\
2-VIS+LSTM: \textcolor{green}{tv (0.29)}\\
LSTM: \textcolor{green}{tv (0.14)}

\vskip 0.05in
DAQUAR 1426a\\
\textbf{What is on the left side of the room?}\\
Ground truth: bed\\
IMG+BOW: \textcolor{red}{door (0.19)}\\
2-VIS+LSTM: \textcolor{red}{door (0.25)}\\
LSTM: \textcolor{red}{door (0.13)}
}
&

\scalebox{0.23}{
\includegraphics[width=\textwidth, height=.7\textwidth]{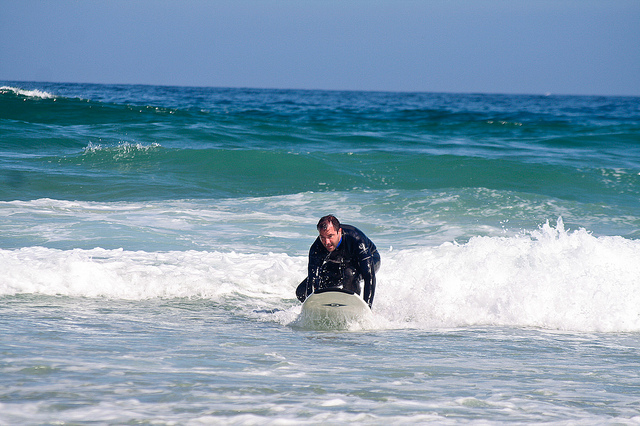}}

\parbox{3.2cm}{
\vskip 0.05in
COCOQA 15756\\
\textbf{What does the man rid while wearing a black wet suit?}\\
Ground truth: surfboard\\
IMG+BOW: \textcolor{red}{jacket (0.35)}\\
2-VIS+LSTM: \textcolor{green}{surfboard (0.53)}\\
BOW: \textcolor{red}{tie (0.30)}
}
&

\scalebox{0.23}{
\includegraphics[width=\textwidth, height=.7\textwidth]{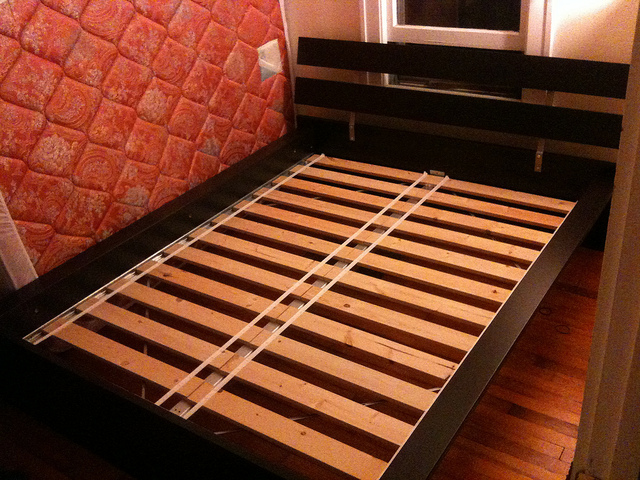}}
\parbox{3.2cm}{
\vskip 0.05in
COCOQA 9715\\
\textbf{What is displayed with the mattress off of it?}\\
Ground truth: bed\\
IMG+BOW: \textcolor{red}{bench (0.36)}\\
2-VIS+LSTM: \textcolor{green}{bed (0.18)}\\
BOW: \textcolor{red}{airplane (0.08)}
}
&

\scalebox{0.23}{
\includegraphics[width=\textwidth, height=.7\textwidth]{25124.jpg}}
\parbox{3.2cm}{
\vskip 0.05in
COCOQA 25124\\
\textbf{What is sitting in a sink in the rest room?}\\
Ground truth: cat\\
IMG+BOW: \textcolor{red}{toilet (0.77)}\\
2-VIS+LSTM: \textcolor{red}{toilet (0.90)}\\
BOW: \textcolor{green}{cat (0.83)}
}
\end{array}$
\caption{Sample questions and responses on Object questions}
\label{fig:object_questions}
\end{figure}

\begin{figure}[ht]
\centering
\tiny
$\begin{array}{p{3.2cm} p{3.2cm} p{3.2cm} p{3.2cm}}
\scalebox{0.23}{
\includegraphics[width=\textwidth, height=.7\textwidth]{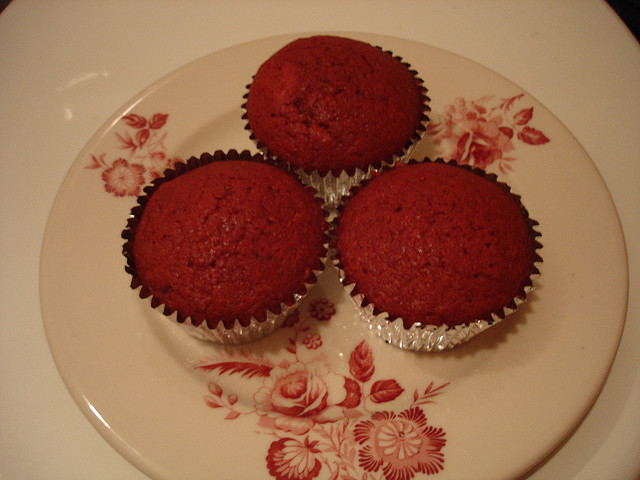}}
\parbox{3.2cm}{
\vskip 0.05in
COCOQA 35\\
\textbf{How many red velvet cup cakes with no frosting on a flowered
plate?}\\
Ground truth: three\\
IMG+BOW: \textcolor{green}{three (0.43)}\\
2-VIS+BLSTM: \textcolor{green}{three (0.26)}\\
BOW: \textcolor{red}{two (0.26)}
}
&

\scalebox{0.23}{
\includegraphics[width=\textwidth, height=.7\textwidth]{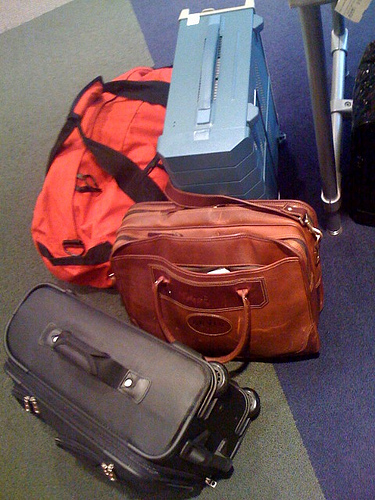}}

\parbox{3.2cm}{
\vskip 0.05in
COCOQA 21446\\
\textbf{How many different types of carrying cases in various
colors?}\\
Ground truth: four\\
IMG+BOW: \textcolor{green}{four (0.69)}\\
2-VIS+BLSTM: \textcolor{green}{four (0.26)}\\
BOW: \textcolor{red}{three (0.25)}
}
&

\scalebox{0.23}{
\includegraphics[width=\textwidth, height=.7\textwidth]{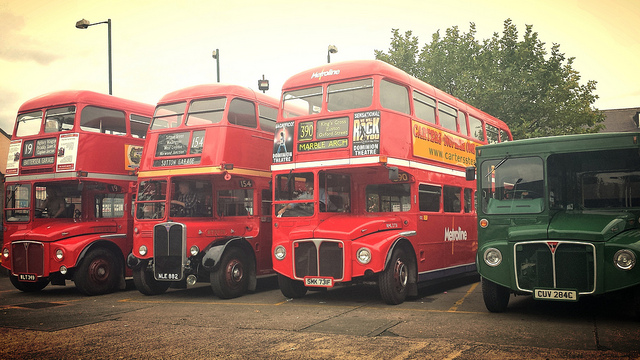}}

\parbox{3.2cm}{
\vskip 0.05in
COCOQA 25600\\
\textbf{How many double deckered busses parked near the green
truck?}\\
Ground truth: three\\
IMG+BOW: \textcolor{red}{two (0.89)}\\
2-VIS+BLSTM: \textcolor{red}{two (0.44)}\\
BOW: \textcolor{red}{two (0.64)}
}
&

\scalebox{0.23}{
\includegraphics[width=\textwidth, height=.7\textwidth]{555.jpg}}
\parbox{3.2cm}{
\vskip 0.05in
DAQUAR 555\\
\textbf{How many chairs are there?}\\
Ground truth: one\\
IMG+BOW: \textcolor{red}{six (0.20)}\\
2-VIS+BLSTM: \textcolor{green}{one (0.20)}\\
LSTM: \textcolor{red}{four (0.19)}
}
\end{array}$

\caption{Sample questions and responses on Number questions}
\label{fig:number_questions}
\end{figure}

\begin{figure}[ht!]
\centering
\tiny
$\begin{array}{p{3.2cm} p{3.2cm} p{3.2cm} p{3.2cm}}

\scalebox{0.23}{
\includegraphics[width=\textwidth, height=.7\textwidth]{16330.jpg}}

\parbox{3.2cm}{
\vskip 0.05in
COCOQA 16330\\
\textbf{What is the color of the bowl?}\\
Ground truth: blue\\
IMG+BOW: \textcolor{green}{blue (0.83) }\\
2-VIS+LSTM: \textcolor{green}{blue (0.90) }\\
BOW: \textcolor{red}{white (0.45}

\vskip 0.05in
COCOQA 16330a\\
\textbf{What is the color of the cat?}\\
Ground truth: gray\\
IMG+BOW: \textcolor{green}{gray (0.45)}\\
2-VIS+BLSTM: \textcolor{red}{blue (0.32)}\\
BOW: \textcolor{red}{gray (0.40)}
}
&

\scalebox{0.23}{
\includegraphics[width=\textwidth, height=.7\textwidth]{2989.jpg}}

\parbox{3.2cm}{
\vskip 0.05in
DAQUAR 2989\\
\textbf{What is the color of the sofa?}\\
Ground truth: red\\
IMG+BOW: \textcolor{green}{red (0.31) }\\
VIS+LSTM-2: \textcolor{green}{red (0.22) }\\
LSTM: \textcolor{red}{brown (0.21) }

\vskip 0.05in
DAQUAR 2989a\\
\textbf{What is the color of the table?}\\
Ground truth: white\\
IMG+BOW: \textcolor{green}{white (0.29) }\\
VIS+LSTM-2: \textcolor{green}{white (0.41) }\\
LSTM: \textcolor{green}{white (0.28) }
}
&

\scalebox{0.23}{
\includegraphics[width=\textwidth, height=.7\textwidth]{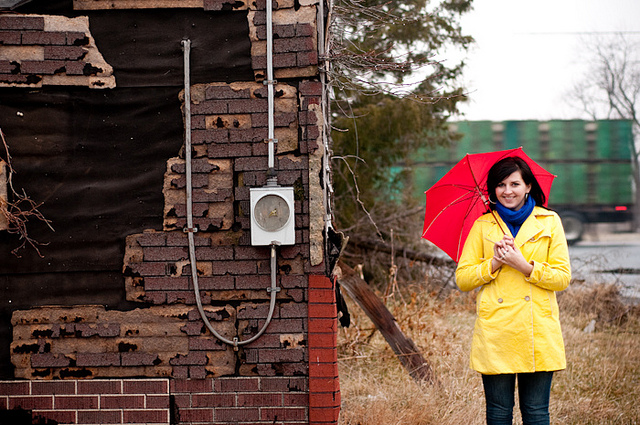}}

\parbox{3.2cm}{
\vskip 0.05in
COCOQA 22891\\
\textbf{What is the color of the coat?}\\
Ground truth: yellow\\
IMG+BOW: \textcolor{red}{black(0.45)}\\
VIS+LSTM: \textcolor{green}{yellow (0.24)}\\
BOW:  \textcolor{red}{red (0.28)}

\vskip 0.05in
COCOQA 22891a\\
\textbf{What is the color of the umbrella?}\\
Ground truth: red\\
IMG+BOW: \textcolor{red}{black(0.28)}\\
VIS+LSTM-2: \textcolor{red}{yellow (0.26)}\\
BOW:  \textcolor{green}{red (0.29) }
}
&
\scalebox{0.23}{
\includegraphics[width=\textwidth, height=.7\textwidth]{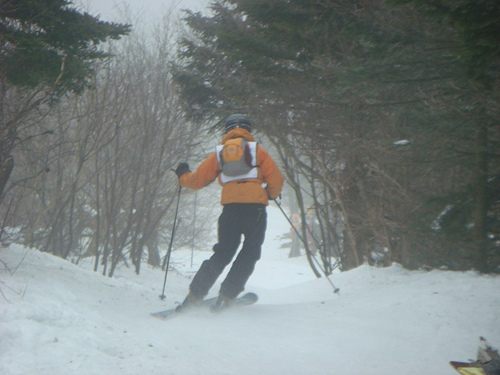}}

\parbox{3.2cm}{
\vskip 0.05in
COCOQA 20140\\
\textbf{What is the color of the coat?}\\
Ground truth: orange\\
IMG+BOW: \textcolor{green}{orange(0.21)}\\
VIS+LSTM: \textcolor{red}{yellow (0.21)}\\
BOW:  \textcolor{red}{red (0.18)}
\vskip 0.05in
COCOQA 20140a\\
\textbf{What is the color of the pants?}\\
Ground truth: black\\
IMG+BOW: \textcolor{green}{black(0.27)}\\
VIS+LSTM-2: \textcolor{green}{black (0.23)}\\
BOW:  \textcolor{green}{black (0.30) }
}
\end{array}$

\caption{Sample questions and responses on Color questions}
\label{fig:color_questions}
\end{figure}

\begin{figure}[ht!]
\centering
\tiny
$\begin{array}{p{3.2cm} p{3.2cm} p{3.2cm} p{3.2cm}}
\scalebox{0.23}{

\includegraphics[width=\textwidth, height=.7\textwidth]{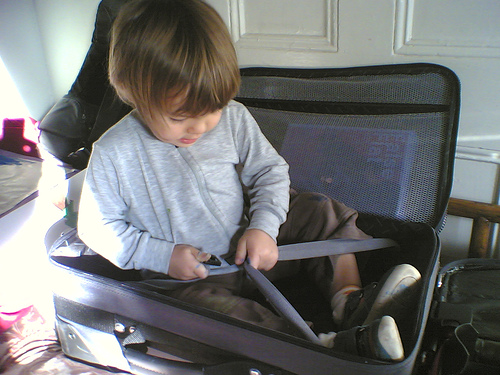}}
\parbox{3.2cm}{
\vskip 0.05in
COCOQA 6427\\
\textbf{Where does the young boy secure himself?}\\
Ground truth: suitcase\\
IMG+BOW: \textcolor{green}{suitcase (0.39)}\\
2-VIS+LSTM: \textcolor{red}{car (0.32)}\\
BOW: \textcolor{red}{mirror (0.23)}
}
&

\scalebox{0.23}{
\includegraphics[width=\textwidth, height=.7\textwidth]{13340.jpg}}

\parbox{3.2cm}{
\vskip 0.05in
COCOQA 13340\\
\textbf{Where are beans and green vegetables?}\\
Ground truth: pan\\
IMG+BOW: \textcolor{green}{pan (0.55)}\\
2-VIS+LSTM: \textcolor{green}{pan (0.44)}\\
BOW: \textcolor{red}{bowl (0.20)}
}
&

\scalebox{0.23}{
\includegraphics[width=\textwidth, height=.7\textwidth]{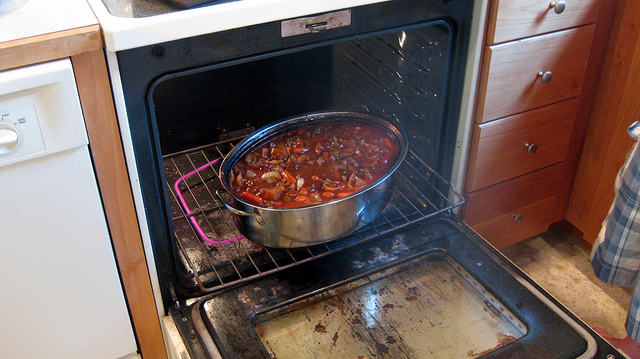}}

\parbox{3.2cm}{
\vskip 0.05in
COCOQA 28542\\
\textbf{Where is the pan of food?}\\
Ground truth: oven\\
IMG+BOW: \textcolor{green}{oven (0.67)}\\
2-VIS+LSTM: \textcolor{green}{oven (0.93)}\\
BOW: \textcolor{red}{kitchen (0.37)}
}
&

\scalebox{0.23}{
\includegraphics[width=\textwidth, height=.7\textwidth]{21112.jpg}}

\parbox{3.2cm}{
\vskip 0.05in
COCOQA 21112\\
\textbf{Where is the cat observing the dishwasher?}\\
Ground truth: kitchen\\
IMG+BOW: \textcolor{red}{sink (0.39)}\\
2-VIS+LSTM: \textcolor{red}{chair (0.44)}\\
BOW: \textcolor{red}{sink (0.17)}
}

\end{array}$
\caption{Sample questions and responses on Location questions}
\label{fig:location_questions}
\end{figure}

\end{document}